\documentclass{article}
\usepackage{ijcai20}

\usepackage[dvipsnames]{xcolor}
\usepackage{times}
\usepackage{soul}
\usepackage{url}
\usepackage[hidelinks]{hyperref}
\usepackage[utf8]{inputenc}
\usepackage[small]{caption}
\usepackage{graphicx}
\usepackage{amsmath}
\usepackage{amsthm}
\usepackage{tikz}
\usepackage{booktabs}
\usepackage{algorithm}
\usepackage{algorithmic}

\usepackage{todonotes}  

\title{Distill, Adapt, Distill: \\
Training Small, In-Domain Models for Neural Machine Translation}

\author{
Mitchell A. Gordon\and
Kevin Duh\\
\affiliations
Johns Hopkins University\\
\emails
\{mitchg, kduh\}@jhu.edu
}

\begin{document}

\maketitle

\begin{abstract}
We explore best practices for training small, memory efficient machine translation models with sequence-level knowledge distillation in the domain adaptation setting. While both domain adaptation and knowledge distillation are widely-used, their interaction remains little understood. Our large-scale empirical results in machine translation (on three language pairs with three domains each) suggest distilling twice for best performance: once using general-domain data and again using in-domain data with an adapted teacher.
\end{abstract}

\section{Introduction}
\begin{figure*}[ht!]
\centering
\begin{tikzpicture}
    \node[rectangle, inner sep=10pt, draw, text width=1cm] at (-1, 1) (GD Teacher){GD Teacher};
    \node[rectangle, inner sep=5pt, draw] at (-1, -2) (GD Baseline){GD Baseline}; 
    \node[rectangle, inner sep=5pt, draw] at (-1, -3) (GD Student){GD Student}; 
    
    \node[rectangle, inner sep=10pt, draw, text width=1cm] at (3.5, 1) (Baseline Teacher) {Baseline Teacher};
    \node[rectangle, inner sep=10pt, draw, text width=1cm] at (5.5, 1) (Adapted Teacher) {Adapted Teacher};
    
    \node[rectangle, inner sep=5pt, draw] at (1.5, -1) (1){1}; 
    \node[rectangle, inner sep=5pt, draw] at (3.5, -1) (2){2}; 
    \node[rectangle, inner sep=5pt, draw] at (5.5, -1) (3){3}; 
    \node[rectangle, inner sep=5pt, draw] at (1.5, -2) (4){4}; 
    \node[rectangle, inner sep=5pt, draw] at (3.5, -2) (5){5}; 
    \node[rectangle, inner sep=5pt, draw] at (5.5, -2) (6){6}; 
    \node[rectangle, inner sep=5pt, draw] at (1.5, -3) (7){7}; 
    \node[rectangle, inner sep=5pt, draw] at (3.5, -3) (8){8}; 
    \node[rectangle, inner sep=5pt, draw] at (5.5, -3) (9){9}; 
    
    \draw[dotted] (0.5, 1.7) -- (0.5, -3.4);
    \node at (1.5, 1) {In-Domain};

    \draw[->] (GD Teacher) .. controls (1.5, 2) and (5.5, 2) .. (Adapted Teacher.north);
    \draw[->, dashed] (GD Teacher.west) .. controls (-3, -1.5) and (-3,-3) .. (GD Student.west);
    
    \draw[->, dashed] (Baseline Teacher) -- (2.north);
    \draw[->, dashed] (Baseline Teacher.south) .. controls (4.5, -1) and (4.5, -2) .. (5.east);
    \draw[->, dashed] (Baseline Teacher.south) .. controls (4.5, -1) and (4.5, -2) .. (8.east);
    
    \draw[->, dashed] (Adapted Teacher) -- (3.north);
    \draw[->, dashed] (Adapted Teacher.south) .. controls (6.5, -1) and (6.5, -2) .. (6.east);
    \draw[->, dashed] (Adapted Teacher.south) .. controls (6.5, -1) and (6.5, -2) .. (9.east);

    \draw[->] (GD Baseline) -- (4.west);
    \draw[->] (3.5, -2.5) -- (5.south);
    \draw[->] (GD Baseline.east) .. controls (1.5, -2.5) .. (3.5, -2.5) -- (5.5, -2.5) -- (6.south);
    
    \draw[->] (GD Student) -- (7.west);
    \draw[->] (3.5, -3.5) -- (8.south);
    \draw[->] (GD Student.east) .. controls (1.5, -3.5) .. (3.5, -3.5) -- (5.5, -3.5) -- (9.south);
    
\end{tikzpicture}
\caption{There are 9 possible configurations for training small, in-domain models with knowledge distillation and domain adaptation. Models trained on general-domain data are shown on the left, and in-domain models are shown on the right. Solid arrows represent domain adaptation via continued training. Dashed arrows represent improved optimization via sequence-level knowledge distillation. Configuration 1 is the model which is trained  on in-domain data with random initializations and without the assistance of a teacher. }
\label{fig:configs}
\end{figure*}
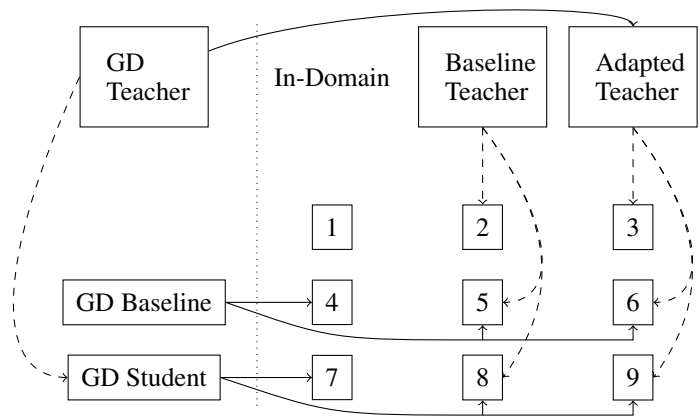

Machine translation systems rely on large amounts of data to deduce the rules underlying translation from one language to another. This presents challenges in some important niche domains, such as patent and medical literature translation, due to the high cost of hiring experts to generate suitable training data. A cost-effective alternative is \emph{domain adaptation}, which leverages large amounts of parallel documents from less difficult and readily-available domains, such as movie subtitles and news articles.

Domain adaptation works well in practice. However, these large datasets, which we call \emph{general domain} datasets, introduce some scalability problems. Large datasets require large models; neural machine translation systems can take days or weeks to train. Some models require gigabytes of disk space, making deployment to edge computing devices challenging. They can also require excessive compute during inference, making them slow and costly to scale up in production environments.\cite{gordon_2019}

To alleviate these issues, \emph{knowledge distillation} (aka Teacher-Student) \cite{Hinton2015-on} is used to compress models into a manageable form. But although knowledge distillation is the most commonly used form of model compression in practice, it is also one of the least understood.

In this work, we perform a \textbf{large-scale empirical analysis} to attempt to discover best practices when using knowledge distillation in combination with domain adaptation. Out of several common-sense configurations, we find that two stages of knowledge distillation give the best performance: one using general-domain data and another using in-domain data with an adapted teacher. We perform experiments on multiple language pairs (Russian-English, German-English, Chinese-English), domains (patents, subtitles, news, TED talks), and student sizes.

\section{Background}

\paragraph{Domain Adaptation} helps overcome a lack of quality training data in niche domains by leveraging large amounts of data in a more accessible general-domain. Domain adaptation is usually accomplished by \emph{continued training} \cite{Luong2015-ym,Zoph2016-pt}, which involves two steps:

\begin{enumerate}
    \item A model is randomly initialized and trained until convergence on the general-domain data.
    \item A new model is initialized with the parameters resulting from Step 1 and trained until convergence on the in-domain dataset.
\end{enumerate}

We can consider domain adaptation as extracting a useful \emph{inductive-bias} from the general-domain dataset, which is encoded and passed along to the in-domain model as a favorable weight initialization. While there are other methods of extracting inductive bias from general-domain datasets (including mixed fine-tuning \cite{Chu2017-db} and cost weighting \cite{Chen2017-vw}), continued training is most common and the focus of this paper. 

\paragraph{Knowledge Distillation} is a method for improving the performance of under-parameterized ``Student" models by exploiting the probability distribution of a more computationally complex ``Teacher" network. Kim and Rush \shortcite{Kim2016-st} presented an extension of knowledge distillation to machine translation in two flavors: word-level and sequence-level knowledge distillation.

 \emph{Sequence-level knowledge distillation}, which is more general, involves three steps:
 
 \begin{enumerate}
     \item A large Teacher network is randomly initialized and trained until convergence on the data.
     \item The source-side of the training data is decoded using the Teacher to produce ``distilled" target data.
     \item A smaller Student model is randomly initialized and trained until convergence on the distilled source-target pairs (discarding the original target sequences in the data).
 \end{enumerate}

The goal of knowledge distillation is to train the student model to mimic the teacher's probability distribution over translations. Since the teacher and the student are trained on the same dataset, they should be capable of learning the same distribution in theory. In practice, however, using the teacher as an additional training signal improves student test performance.\footnote{Interestingly, this can be true even when the student has the same computational resources as the teacher \cite{Furlanello2018-lp}} Explanations for this phenomenon include dark knowledge \cite{Furlanello2018-lp}, mode reduction \cite{Zhou2019-wp}, and regularization \cite{Gordon2019-qs,Dong2019-ve}, but no definitive evidence has been given.

 Sequence-level knowledge distillation is widely used in both industry \cite{Xia2019-iq} and research and is the second focus of this paper. \footnote{Sequence-level knowledge distillation is also commonly used to train non-autoregressive machine translation models \cite{Zhou2019-wp}.}

\section{Distilling and Adapting}

How domain adaptation and knowledge distillation would interact when applied in combination was not previously clear. Specifically, our research questions are:

\begin{itemize}
    \item Is a distilled model easier or harder to adapt to new domains?
    \item Should knowledge distillation be used on in-domain data? If so, how should the teacher be trained?
\end{itemize}

To answer these questions, we performed experiments on 9 possible configurations which are assigned configuration numbers in Figure \ref{fig:configs}. For ease of reference, we will primarily refer to small, in-domain models by their configuration number and encourage readers to consult Figure \ref{fig:configs}. Each configuration has two attributes of interest. 

\paragraph{Distilling In-Domain Data} How is in-domain data pre-processed using knowledge distillation? Some models are trained with no pre-processing (configurations 1, 4, and 7), while others use a teacher to pre-process the in-domain training data. This teacher might be a baseline trained on in-domain data only (configurations 2, 5, and 8) or it can be trained on general-domain data and then adapted to in-domain via continued training (configurations 3, 6, and 9).

\paragraph{Initialization} How are models initialized? A model might be randomly initialized (configurations 1, 2, and 3), or it might be adapted from a model trained on general-domain data. This general-domain model might be a baseline trained directly on the general-domain data (configurations 4, 5, and 6) or it might be a student model trained on the output of a general-domain teacher (configurations 7, 8, 9).

\section{Experiments}
\subsection{Data}
\paragraph{General-Domain Data}
We train models in multiple settings: 3 language pairs (German-English, Russian-English, and Chinese-English) each with 1 general-domain dataset and 2 different in-domain datasets. 
The general-domain datasets for each language are a concatenation of data from OpenSubtitles2018 
\cite{Tiedemann2016-ob,Lison2016-kr} (which contains translated movie subtitles) and the WMT 2017 datasets \cite{Ondrej2017-dw} (which includes a variety of sources, including news commentary, parliamentary proceedings, and web-crawled data).

\paragraph{In-Domain Data}
We use the World International Property Organization (WIPO) COPPA-V2 dataset \cite{Junczys-Dowmunt2018-oh} and the TED Talks dataset \cite{Qi2018-le} as our two in-domain datasets. The WIPO data contains parallel sentences from international patent abstracts, while the TED Talks dataset consists of translated transcripts of public speeches.

\paragraph{Data Statistics} The size of each training dataset is presented in Table \ref{tab:data-sizes}. General-domain datasets contain tens of millions of sentences, while in-domain datasets contain much less. German-English WIPO has an exceptional amount of training data (4.5 times more than the next biggest in-domain dataset) and helps qualify how our results might change when more in-domain data is available.

\begin{table}[]
    \centering
    \begin{tabular}{lrrr}
        Language & General-Domain & WIPO & TED\\
        \hline
        De-En & 28.3 M & 821 k & 15 k\\
        Ru-En & 51.1 M & 29 k & 180 k\\
        Zh-En & 35.9 M & 154 k & 169 k\\
        
    \end{tabular}
    \caption{The number of training sentences in each dataset.}
    \label{tab:data-sizes}
\end{table}

\paragraph{Pre-processing} All datasets are tokenized using the Moses\footnote{\href{http://statmt.org/moses}{statmt.org/moses}} tokenizer. A BPE vocabulary \cite{Sennrich2016-an} of 30,000 tokens is constructed for each language using the training set of the general-domain data. This BPE vocabulary is then applied to both in-domain and general-domain datasets. This mimics the typical scenario of a single, general-domain model being trained and then adapted to new domains as they are encountered. Note that re-training BPE on in-domain data to produce a different vocabulary would force us to re-build the model, making adaptation impossible.

\paragraph{Evaluation} The general-domain development set for each language contains newstest2016 concatenated with the last 2500 lines of OpenSubtitles2018. We reserve 3000 lines of WIPO to use as the in-domain development set. TED talks development sets are provided by the authors and contain around 2000 lines each. Evaluations of each model are performed by decoding the appropriate development set with a beam-search size of 10 and comparing to the reference using multi-bleu.perl from the Moses toolkit.

\subsection{Architectures and Training}
 A list of architecture sizes is provided in Table \ref{tab:sizes}. Teachers are trained using the Large hyper-parameter settings, while we experiment with Medium, Small, and Tiny students for each configuration and language/domain setting.
 
 All models are Transformers \cite{Vaswani2017-vu}. We use the same hyper-parameters (which are based on a template from \cite{Duh_undated-uo}\footnote{\href{https://git.io/JvL85}{https://git.io/JvL85}}) for every model, except those that affect the size of the model (Table \ref{tab:sizes}). Models are trained either for 300,000 updates, 100 epochs, or until the model does not improve for 10 checkpoints (early-stopping), whichever comes first.
 
 \begin{table}[]
     \centering
     \begin{tabular}{lccc}
          Size & Layers & FF Size & Hidden Size \\
          \hline
          Large & 12 & 2048 & 512 \\
          Medium & 6 & 2048 & 512 \\
          Small & 6 & 1024 & 256 \\
          Tiny & 2 & 1024 & 256 \\
     \end{tabular}
     \caption{Hyper-parameters of various model sizes used in this work. For example, the Large Transformer model architecture uses 6 encoder and 6 decoder layers, a feed-forward hidden dimension of 2048 at each layer, and a word-embedding / hidden dimension of 512.}
     \label{tab:sizes}
 \end{table}

\paragraph{Continued Training} Work by \cite{Gordon2019-qs} suggests that students may benefit from training on some combination of the distilled and un-distilled reference dataset. We experimented with this by continuing to train each in-domain student model on the original, un-distilled dataset, using similar stopping criterion to the first round of training. This improved some models by up to 1 BLEU. Because of this, we recommend that any distilled model continue training on the original dataset as long as development accuracy improves. When continued training improves performance of a student, we show that score instead of the score without continued training.

\section{Recommendations}
\subsection{Adapt Teachers}
    
In this section, we compare training in-domain models with no teacher (config 1), a teacher trained on in-domain data only (config 2), and a teacher adapted from the general domain (config 3). The performance of the two teachers in each language-pair and domain is listed in Table \ref{tab:id-teachers}. It shows that adaptation greatly improves the performance of every in-domain teacher except German-English WIPO.\footnote{German WIPO is also the largest in-domain dataset we test, which might make adaptation unnecessary. Another explanation might be that the German-English general-domain is not similar enough to the patent domain in this case to improve performance.}

Table \ref{tab:id-students} shows the results of using these teachers to distill the in-domain data before training student models in various settings. \textbf{We see that in almost every case, using an adapted teacher gives the best or close to the best results.} This is somewhat expected since models with better development scores tend to make better teachers \cite{Zhang-2018ak}. Although knowledge distillation is typically seen as ``simplifying" data for students, in this case we suspect that the adapted teacher's knowledge about the general-domain is making its way to students via the distilled in-domain data.
    \begin{table}
    \centering
    \begin{tabular}{lllccc}
    Domain & Size & Adapted From & de-en & ru-en & zh-en\\
    
\hline
ted & Large & None                       & 29.25 & 19.38 & 14.79\\
 &  & Large                      & \color{ForestGreen}37.64 & \color{ForestGreen}26.57 & \color{ForestGreen}20.45\\
\hline
wipo & Large & None                      & 48.31 & 21.36 & 31.02\\
 &  & Large                     & \color{ForestGreen} 48.56 & \color{ForestGreen}37.08 & \color{ForestGreen}36.80\\

    \end{tabular}
    \caption{BLEU development score of in-domain teachers. Adaptation drastically improves performance on every language pair and domain, except de-en WIPO.}
    \label{tab:id-teachers}
    \end{table}

    \begin{table}
    \centering
    \begin{tabular}{llcccc}
    Domain & Size & Config \# & de-en & ru-en & zh-en\\
    
 &  & 1                                  & 27.73 & 19.34 & 15.17\\
ted & medium & 2                         & 29.11 & 20.31 & 15.71\\
 &  & 3                                  & \textbf{29.54} & \textbf{20.56} & \textbf{15.90}\\
\hline
 &  & 1                                  & 27.89 & 18.42 & 14.87\\
 & small & 2                             & 28.93 & 19.65 & 14.95\\
 &  & 3                                  & \textbf{29.52} & \textbf{19.88} & \textbf{15.79}\\
\hline
 &  & 1                                  & 25.78 & 17.48 & 13.03\\
 & tiny & 2                              & 27.20 & 17.87 & 13.39\\
 &  & 3                                  & \textbf{27.58} & \textbf{19.27} & \textbf{13.74}\\
\hline
 &  & 1                                  & 48.89 & 24.45 & 30.13\\
wipo & medium & 2                        & \textbf{50.66} & \textbf{24.62} & 32.13\\
 &  & 3                                  & 50.23 & 24.60 & \textbf{33.16}\\
\hline
 &  & 1                                  & 47.94 & 21.91 & 30.66\\
 & small & 2                             & 49.46 & \textbf{23.70} & 32.19\\
 &  & 3                                  & \textbf{49.72} & 23.50 & \textbf{32.61}\\
\hline
 &  & 1                                  & 44.15 & 21.39 & 27.67\\
 & tiny & 2                              & 48.03 & \textbf{22.24} & 28.18\\
 &  & 3                                  & \textbf{48.51} & 22.03 & \textbf{29.88}\\

    \end{tabular}
    \caption{BLEU development scores for in-domain students with no teacher (config 1), an in-domain only teacher (config 2), or an adapted teacher continued from the general-domain (config 3). In almost every case, using an adapted teacher gives the best or close to the best results. }
    \label{tab:id-students}
    \end{table}

\subsection{Adapt the Best Student}

We also train small models directly on the general-domain data and adapt them to in-domain data. The possible configurations are random initialization (config 1), initializing from a baseline model trained on general-domain data (config 4), or initializing from a student model distilled from a general-domain teacher (config 7). Table \ref{tab:gd} shows the performance of these models on the general-domain datasets, and Table \ref{tab:id-adapted} shows their performance on in-domain datasets.

While adapting teachers gives modest gains on in-domain datasets (0-2 BLEU), training small models directly on the general-domain data gives much more substantial gains (5-10 BLEU). We believe this is because a large amount of data is required to fully reveal the teacher's probability distribution over translations \cite{Fang2019-lw}. While an adapted teacher might contain much information from the general-domain, it is unable to express that knowledge to students just by translating the smaller in-domain dataset. \textbf{To get the full benefit of general-domain data, the small models must be directly trained on general-domain data.}\footnote{A reasonable alternative to this might include data-free KD \cite{Yin2019-bm}, which explores the teacher's probability distribution without any dependence on data.}

\begin{figure}
        \centering
        \includegraphics[width=9cm]{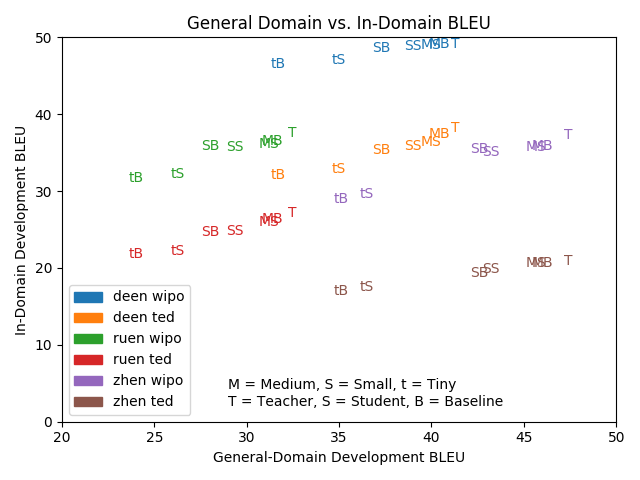}
        \caption{The BLEU of general-domain models vs. their corresponding in-domain scores when adapted to a different domain. We see that in-domain performance is roughly linearly related to general-domain performance regardless of whether distillation is applied before adaptation.}
        \label{fig:gd-id}
    \end{figure}

We also observe that Medium-sized models are not small enough to benefit from knowledge distillation in the general-domain, and so their general-domain scores do not improve with distillation. These distilled Medium-sized models (config 7) also tend to do slightly worse than their baseline counter-parts (config 4) on in-domain data. Indeed, Figure \ref{fig:gd-id} shows that in-domain performance is roughly linearly related to general-domain performance regardless of whether distillation is applied before adaptation.

This implies that \textbf{distillation does not interfere with the adaptability of a model}, so the model with the best general-domain performance should be adapted, regardless of whether distillation was applied. Adapting a distilled model can improve performance by up to 1 BLEU over adapting the baseline model without distillation.

    \begin{table}
    \centering
    \begin{tabular}{lccc}
    Model & de-en & ru-en & zh-en\\
    
\hline
Teacher                                  & 41.08 & 32.25 & 47.17\\
\hline
Medium Baseline                          & 39.86 & 30.81 & 45.40\\
Medium Student                           &\color{red} 39.40 & \color{red} 30.65 & \color{red} 45.11\\
\hline
Small Baseline                           & 36.78 & 27.54 & 42.09\\
Small Student                            & \color{ForestGreen} 38.51 & \color{ForestGreen}28.88 & \color{ForestGreen}42.73\\
\hline
Tiny Baseline                            & 31.27 & 23.63 & 34.71\\
Tiny Student                             & \color{ForestGreen}34.58 & \color{ForestGreen}25.86 & \color{ForestGreen}36.09\\

    \end{tabular}
    \caption{General-domain models, teachers and students. While knowledge distillation improves small and tiny models, it appears medium-sized models are not under-parameterized enough for knowledge distillation to improve performance.}
    \label{tab:gd}
    \end{table}

    \begin{table}
    \centering
    \begin{tabular}{llcccc}
    Domain & Size & Config \# & de-en & ru-en & zh-en\\

\hline
 &  & 1                                  & 27.73 & 19.34 & 15.17\\
ted & medium & 4                         & \textbf{36.94} & \textbf{25.82} & 20.13\\
 &  & 7                                  & 35.93 & 25.43 & \textbf{20.18}\\
\hline
 &  & 1                                  & 27.89 & 18.42 & 14.87\\
 & small & 4                          & 34.78 & 24.10 & 18.84\\
 &  & 7                                  & \textbf{35.33} & \textbf{24.30} & \textbf{19.32}\\
\hline
 &  & 1                                  & 25.78 & 17.48 & 13.03\\
 & tiny & 4                           & 31.52 & 21.30 & 16.51\\
 &  & 7                                  & \textbf{32.30} & \textbf{21.65} & \textbf{17.06}\\
\hline
 &  & 1                                  & \textbf{48.89} & 24.45 & 30.13\\
wipo & medium & 4                        & 48.58 & \textbf{35.98} & \textbf{35.33}\\
 &  & 7                                  & 48.53 & 35.55 & 35.27\\
\hline
 &  & 1                                  & 47.94 & 21.91 & 30.66\\
 & small & 4                         & 48.13 & \textbf{35.30} & \textbf{34.90}\\
 &  & 7                                  & \textbf{48.31} & 35.18 & 34.52\\
\hline
 &  & 1                                  & 44.15 & 21.39 & 27.67\\
 & tiny & 4                          & 46.06 & 31.13 & 28.45\\
 &  & 7                                  & \textbf{46.54} & \textbf{31.74} & \textbf{29.07}\\

    \end{tabular}
    \caption{In-domain models that are initialized randomly (config 1), initialized from a baseline trained on general-domain data directly (config 4), or initialized from a  general-domain student trained using a general-domain teacher (config 7). }
    \label{tab:id-adapted}
    \end{table}
    
\subsection{Distill, Adapt, Distill}
Finally, we test whether these two ways of improving small, in-domain models are orthogonal. We might hypothesize that training small models directly on general-domain data eliminates the need to adapt teachers or use an in-domain teacher at all. To test this, we also train adapted student models using a baseline teacher (config 8) and an adapted teacher (config 9).

   \begin{table}
    \centering
    \begin{tabular}{llcccc}
    Domain & Size & Config \# & de-en & ru-en & zh-en\\
    
\hline
 &  & 7                                  & 35.93 & 25.43 & \textbf{20.18}\\
ted & medium & 8                         & 35.23 & 25.18 & 19.96\\
 &  & 9                                  & \textbf{36.65} & \textbf{25.91} & 20.13\\
\hline
 &  & 7                                  & 35.33 & 24.30 & 19.32\\
 & small & 8                             & 35.11 & 23.97 & 19.17\\
 &  & 9                                  & \textbf{35.57} & \textbf{24.95} & \textbf{19.48}\\
\hline
 &  & 7                                  & 32.30 & 21.65 & 17.06\\
 & tiny & 8                              & 32.21 & 21.45 & 16.72\\
 &  & 9                                  & \textbf{33.12} &\textbf{22.49} & \textbf{17.54}\\
\hline
 &  & 7                                  & 48.53 & 35.55 & 35.27\\
wipo & medium & 8                        & 49.07 & 34.71 & 35.09\\
 &  & 9                                  & \textbf{49.82} & \textbf{35.83} & \textbf{36.48}\\
\hline
 &  & 7                                  & 48.31 & \textbf{35.18} & 34.52\\
 & small & 8                             & \textbf{48.79} & 34.27 & 34.89\\
 &  & 9                                  & 48.35 & 35.10 & \textbf{35.55}\\
\hline
 &  & 7                                  & 46.54 & 31.74 & 29.07\\
 & tiny & 8                              & \textbf{49.90} & 31.12 & 30.05\\
 &  & 9                                  & 49.70 & \textbf{31.75} & \textbf{31.82}\\

    \end{tabular}
    \caption{In-domain models which are initialized from a general-domain student and trained on in-domain data which is pre-processed either with no teacher (config 7), an in-domain only teacher (config 8), or an adapted teacher continued from general-domain data (config 9).}
    \label{tab:id-adapt-both}
    \end{table}

Table \ref{tab:id-adapt-both} shows that \textbf{distilling a second time using in-domain data with an adapted teacher can further boost performance of an already distilled model}, while using an un-adapted in-domain teacher can sometimes hurt performance. 

These results lead us to a general recipe for training small, in-domain models using knowledge distillation and domain adaptation in combination:

\begin{enumerate}
    \item Distill general-domain data to improve general-domain student performance.
    \item Adapt the best model from Step 1 to in-domain data.\\
    (2-10 BLEU better than no adaptation)
    \item Adapt the teacher and distill again in-domain. \\
    (0-2 BLEU better than no or non-adapted teacher)
\end{enumerate}

Following this procedure will result in either configuration 6 or 9 as described in Figure \ref{fig:configs}. And indeed, configuration 9 performs the best or near best (within 0.1 BLEU) in almost every case, as shown in Table \ref{tab:best-configs}. For those Medium sized models which were not improved by distillation in the general-domain, configuration 6 performs the best.

    \begin{table}
    \centering
    \begin{tabular}{llcccc}
    Domain & Size & Config \# & de-en & ru-en & zh-en\\
    
\hline
 & medium & 6                            & 36.80 & 26.26 & 20.13\\
ted & small & 6                          & 35.50 & 24.68 & 19.31\\
 & tiny & 6                              & 32.09 & 22.20 & 17.25\\
\hline
 & medium & 6                            & 48.31 & 35.82 & 36.58\\
wipo & small & 6                         & 49.04 & 35.30 & 35.40\\
 & tiny & 6                              & 48.02 & 31.57 & 30.53\\

    \end{tabular}
    \caption{Development scores for models initialized from a model trained on general-domain data. The in-domain data is pre-processed with a teacher adapted from the general-domain (config 6).}
    \label{tab:config-6}
    \end{table}

Models trained on German-English WIPO are an exception, with adaptation from the general-domain not improving performance. This is in line with the results from Table \ref{tab:id-teachers} which shows adaptation does not improve teachers, either. We suspect this is because the German-English WIPO dataset is the biggest out of any in-domain dataset, making adaptation unnecessary. Future work might also benefit from a quantification of domain similarity between datasets \cite{Britz2017-qg}, which would guide the use of domain adaptation in cases like these.
    
    \begin{table}
    \centering
    \begin{tabular}{llccc}
    Domain & Size & de-en & ru-en & zh-en\\
    
\hline
 & medium & 4/6 & 6 & 4/6/7/9\\
ted & small & 6/9 & 9 & 9\\
 & tiny & 9 & 9 & 9\\
\hline
 & medium & 2 & 4/6/9 & 6\\
wipo & small & 3 & 4/6 & 9\\
 & tiny & 8 & 7/9 & 9\\

    \end{tabular}
    \caption{Best configurations for each setting. Scores within 0.1 BLEU of the best are also listed. Configuration 9 generally performs best, while configuration 6 is best for those medium-sized models which were not improved by distillation in the general-domain.}
    \label{tab:best-configs}
    \end{table}

\subsection{Training Times}
The models trained in this work collectively required 10 months of single-GPU compute time. Table \ref{tab:times} breaks this down by model size and dataset.

While distilling twice might give the best performance, it also increases the amount of computation time required. Rather than training a single in-domain model, configuration 9 requires training a general-domain teacher, a general-domain student, and then adapting both. This can increase compute required to train models by 2-4x.

A huge portion of computation was also spent on decoding the general-domain data using a teacher model for sequence-level knowledge distillation, which could take up to 24 days of GPU time (using a beam size of 10 and a batch size of 10). This can be arbitrarily sped up using multiple GPUs in parallel, but future work might explore how to distill teachers in a less expensive way.

\begin{table}[]
    \centering
    \begin{tabular}{lccc}
        Model & General-Domain & In-Domain & Adapting \\
        \hline
        Large & 2-4 days & 2-4 days & 7-48 hours \\ 
        Medium & 2-4 days & 2-4 days & 1-48 hours \\ 
        Small & 1-2 days & 1-2 days & 2-14 hours \\ 
        Tiny & 1 days & 1-24 hours & 2-24 hours \\
        \hline
        Distilling & 10-24 days & 1-2 days &
    \end{tabular}
    \caption{Estimates of the computation time required for training randomly initialized models on just general-domain data or just in-domain data. We also show the time required for adapting general-domain models and distilling data using teachers.}
    \label{tab:times}
\end{table}

\section{Related Work}
Our work is one the few that focuses specifically on training small, under-parameterized in-domain models. There is, however, similar work which is \emph{not directly comparable} but uses knowledge distillation to adapt to new domains.

\paragraph{Knowledge Adaptation} uses knowledge distillation to transfer knowledge from multiple, labeled source domains to un-labeled target domains. This is in contrast to our setting, which has labels for both general-domain and in-domain data. Ruder et al. \shortcite{Ruder2017-zt} introduced this idea as ``Knowledge Adaptation," using multi-layer perceptrons to provide sentiment analysis labels for unlabeled in-domain data. Similar work includes Iterative Dual Domain Adaptation \cite{Zeng2019-eg} and Domain Transformation Networks \cite{Wang2019-hs}. These ideas are not limited to machine translation; recent work by Meng et al. \shortcite{Meng2020-yu} trains in-domain speech recognition systems with knowledge distillation, while Orbes-Arteaga et al. \shortcite{Orbes-Arteaga2019-de} does similar work on segmentation of magnetic resonance imaging scans. 

\paragraph{Compressing Pre-trained Language Models} Domain adaptation via continued training in NMT is closely related to the idea of pre-training a language model and fine-tuning to different tasks, which might come from different data distributions than the pre-training data. Because language models tend to be extremely cumbersome to train and evaluate, more focus is given to the compression aspect of knowledge distillation. Sanh et al. \shortcite{Sanh2019-gl}, Sun et al. \shortcite{Sun2019-io}, and Liu et al. \shortcite{Liu2019-ho} independently showed that knowledge distillation could be used to compress pre-trained models without affecting downstream tasks. Tang et al. \shortcite{Tang2019-qc} showed that task-specific information could be distilled from a large Transformer into a much smaller Bi-directional RNN. These methods might reasonably be extended to domain adaptation for NMT.

\section{Conclusion}
In this work, we conducted a large-scale empirical investigation to determine best practices when using sequence-level knowledge distillation and domain adaptation in combination. We found that adapting models from the general-domain makes them better teachers and that distilling using general-domain data does not impact a model's adaptability. This leads us to recommend distilling twice for best results: once in the general-domain to possibly improve student performance, and again using an adapted in-domain teacher. The results are robust among multiple language pairs, student sizes, in-domain settings.
\bibliographystyle{named}
\bibliography{ijcai20}

\end{document}